\PassOptionsToPackage{hyphens}{url}
\documentclass[letterpaper, 10 pt, conference]{ieeeconf}  %
\IEEEoverridecommandlockouts                              %
\usepackage{cite}
\usepackage{amsmath,amssymb,amsfonts}
\usepackage{algorithmic}
\usepackage{graphicx}
\usepackage{textcomp}
\usepackage{xcolor}
\usepackage{color}
\usepackage{textcomp}
\usepackage{tabularx}
\usepackage{multirow}

\usepackage{verbatim}
\usepackage{tablefootnote}

\usepackage{siunitx}
\usepackage{booktabs}  
\usepackage{threeparttable} 
\usepackage{xcolor}
\usepackage{etoolbox}
\usepackage{hyperref}

\def\BibTeX{{\rm B\kern-.05em{\sc i\kern-.025em b}\kern-.08em
    T\kern-.1667em\lower.7ex\hbox{E}\kern-.125emX}}

\begin{document}
\bstctlcite{IEEEexample:BSTcontrol}

\title{\LARGE \bf
Siamese Multiple Attention Temporal Convolution Networks for Human Mobility Signature Identification
}

\author{Zhipeng Zheng, Yuchen Jiang, Shiyao~Zhang,~\IEEEmembership{Member, IEEE},
and Xuetao Wei,~\IEEEmembership{Member,~IEEE} \thanks{The authors are with the Research Institute for Trustworthy Autonomous Systems and Department of Computer Science and Engineering, Southern University of Science and Technology, China. This work is supported in part by the Stable Support Plan Program of Shenzhen Natural Science Fund (No. 20220815111111002), and in part by the Guangdong Key Program (No. 2021QN02X166).
Shiyao Zhang is the corresponding author.}}

\maketitle
\thispagestyle{empty}
\pagestyle{empty}
\begin{abstract}
The Human Mobility Signature Identification (HuMID) problem stands as a fundamental task within the realm of driving style representation, dedicated to discerning latent driving behaviors and preferences from diverse driver trajectories for driver identification. 
Its solutions hold significant implications across various domains (e.g., ride-hailing, insurance), wherein their application serves to safeguard users and mitigate potential fraudulent activities.
Present HuMID solutions often exhibit limitations in adaptability when confronted with lengthy trajectories, consequently incurring substantial computational overhead. 
Furthermore, their inability to effectively extract crucial local information further impedes their performance.
To address this problem, we propose a Siamese Multiple Attention Temporal Convolutional Network (Siamese MA-TCN) to capitalize on the strengths of both TCN architecture and multi-head self-attention, enabling the proficient extraction of both local and long-term dependencies.
Additionally, we devise a novel attention mechanism tailored for the efficient aggregation of multi-scale representations derived from our model. 
Experimental evaluations conducted on two real-world taxi trajectory datasets reveal that our proposed model effectively extracts both local key information and long-term dependencies. 
These findings highlight the model's outstanding generalization capabilities, demonstrating its robustness and adaptability across datasets of varying sizes.

\end{abstract}


\section{Introduction}

With the widespread adoption of GPS devices, a substantial volume of vehicle trajectory data is generated and accumulated on a daily basis.
These data are crucial in triggering a myriad of location-based services (e.g., ride-hailing, navigation, point-of-interest recommendation), thereby fostering the advanced development of intelligent transportation systems.
The extraction and utilization of identifiable information embedded within the trajectory data are recognized as pivotal factors within these services. 
Consequently, one notable downstream task in trajectory identifiable information extraction, known as the Human Mobility Signature Identification (HuMID) problem \cite{st_siamese}, has garnered substantial research attention. 
The HuMID problem is defined as the discernment of whether a set of trajectories originates from a claimed specific driver, predicated upon historical trajectories from various drivers.
HuMID finds extensive application in impostor detection across multiple domains, including fleet management, ride-hailing, and insurance, which aims at safeguarding the security of service users and detecting possible fraud. 
For instance, within ride-hailing services, HuMID serves to detect irregular driving patterns and prevent vehicles from being operated by unauthorized individuals, thereby ensuring passenger safety.

The HuMID problem can be conceptualized as a pivotal task within the domain of driving style representation, wherein the crux lies in discerning latent driving behaviors and preferences from diverse drivers' trajectories to derive distinctive representations. 
Prior research in this domain has already made notable contributions.
For instance, Chowdhury \textit{et al.} \cite{Chowdhury} extracted 137 statistical features from smartphone sensors and used a random forest classifier to classify 4 to 5 drivers with relative accuracy.
However, conventional machine-learning approaches rely heavily on extensive human expertise and intricate feature engineering.
Therefore, based on the deep learning techniques, there are several existing studies focusing on implicitly extracting intrinsic information from trajectories, since Dong \textit{et al.} \cite{dong} first attempted to use deep neural networks to learn driving style features in 2016.

Dong \textit{et al.} \cite{ARNet} proposed a unified architecture combining supervised and unsupervised learning, which enhanced the efficacy of learned driving style representations, particularly for novel drivers not encountered during the training phase
In addition, Kieu \textit{et al.} \cite{kieu} proposed the T2INet, a multi-task deep learning framework, which initially maps trajectories to images based on the map grid, then capturing both geographic and driving behavior features for driver number estimation and driver identification.
However, these studies addressing driving style representation as a multi-classification problem exhibit commendable performance solely with a limited cohort of drivers. 
When confronted with an expansive pool of driver candidates, the performance of these approaches notably deteriorates, rendering them unsuitable for real-world HuMID scenarios.

To address tasks characterized by a high number of classes and a limited number of samples per class (e.g., handwritten signature verification and face recognition), the Siamese network structure, as a prominent architecture in metric learning, is frequently employed. 
Ren \textit{et al.} \cite{st_siamese} initially introduced the siamese network structure for processing spatio-temporal data.
They successfully applied this architecture to the HuMID task, which involved 197 drivers not present in the training phase.
By employing two siamese Long Short-Term Memory (LSTM) networks, they demonstrate the considerable potential of the siamese network structure in addressing large-scale driver identification problems.
However, LSTM, as a variant of recurrent neural networks (RNNs), suffers from sequential computation, where the output of each unit depends on the outputs of its preceding unit.
This characteristic inhibits the exploitation of parallel computing resources, resulting in significant time overheads when processing lengthy trajectories.
Additionally, LSTM struggles to efficiently extract spatial information, making it more suitable for processing time series data rather than spatio-temporal trajectory data.

To address the aforementioned limitations, we proposed the Multiple Attention Temporal Convolutional Networks (MA-TCN) for multi-scale trajectory representation learning.
Augmented with extracted profile features, the Siamese MA-TCN is deployed to discern and characterize the behavioral patterns of an extensive cohort of drivers exclusively relying on GPS trajectory data.
The proposed model with flexible receptive fields and better parallelism, offers distinct advantages in capturing long-term dependencies.
Moreover, the convolution operation's spatial invariance facilitates the extraction of local spatial information from trajectories.
Furthermore, the incorporation of multi-head self-attention, coupled with the proposed multi-scale aggregation attention, attracts more key information across multiple dimensions.
Our main contributions are summarized as follows:
\begin{itemize}
    \item We propose Siamese Multiple Attention Temporal Convolutional Networks (Siamese MA-TCN) for human mobility signature identification, in which Siamese MA-TCN is able to efficiently provide accurate predictions while accounting for large-scale driver identifications based on solely trajectory data.
    \item We implement the multi-head self-attention mechanism for acquiring global dependencies and a specially designed aggregation attention mechanism for capturing multi-scale local features from vehicle trajectories.
    Ablation experiments verified the effectiveness of these components.
    \item We provide comprehensive case studies on two real-world city datasets, and demonstrate the outstanding generalization capabilities of our proposed model.
\end{itemize}

\section{Related Work} \label{sec:relatedwork}

\subsection{Driving Style Time-series Deep Learning}

Numerous previous studies \cite{canbus1, canbus2, siamese_canbus} have been dedicated to the exploration of learning driving styles from time series data employing deep learning methodologies.
However, these investigations predominantly rely on the utilization of multiple sensor data (e.g., accelerator pedal value, steering signals, etc.) collected from CAN-BUS.
These data are not only resource-intensive to procure but also tend to contain redundant information.
However, considering the widespread use of in-vehicle GPS devices, vehicle trajectory data is easier to collect and inherently rich in identifiable information.
Consequently, leveraging vehicle trajectory data presents a promising avenue for driving style characterization and driver identification endeavors.

\subsection{Representation Learning with Driver Identification}

Traditional approaches typically rely on classical machine learning algorithms and complex feature engineering, heavily depending on human expertise and failing to fully exploit latent information and potential features in the data.
The earliest relevant research based on deep learning methods can be traced back to 2016 when Dong \textit{et al.} \cite{dong} first attempted to learn driving style features directly from GPS data using deep learning. 
However, supervised learning methods have a clear limitation: the set of identified drivers must be a subset of available labeled drivers during the training phase, making it impossible to identify drivers not included in the labels. 

To address this drawback, Dong \textit{et al.} \cite{ARNet} and Kieu \textit{et al.} \cite{kieu} have successively proposed deep learning frameworks that combine supervised and unsupervised learning and improve the representation capability of unseen data during the training phase.
Liu \textit{et al.} \cite{radar} further considered multiple contextual information, including road conditions, geographic semantics, and traffic conditions, and utilized a semi-supervised Generative Adversarial Network (GAN) to generate more precise representations of driving styles. 

To solve the problem of insufficient identification accuracy in the case of a large number of classes and few intra-class samples, Ren \textit{et al.} \cite{st_siamese} first introduced the Siamese network structure in the processing of spatio-temporal data for verifying whether paired trajectory data is generated by the same driver.
The advantage of this method lies in its ability to identify newly added drivers without historical data while maintaining high accuracy, showcasing the robust potential of the Siamese network structure in addressing large-scale driver identification challenges.

\subsection{Attention Mechanism on Vehicle Trajectory Data}

Attention mechanism was first proposed by Vaswani \textit{et al.} \cite{vaswani2017attention} in 2017, which proposed a novel way to deal with sequence data by Scaled Dot-Product Attention and Multi-Head Attention. 
Due to its superior ability to model long-term dependencies in time-series data, the attention mechanism is applied in several kinds of vehicle trajectory analysis tasks, including trajectory prediction \cite{attention_trajectory_prediction} \cite{attention_trajectory_prediction2}, traffic forecasting \cite{traffic_forecasting1}, and traffic data imputation \cite{ye2021spatial}. 
This study endeavors to extend their utility to the driver identification domain. 
To this end, we introduce multiple attention mechanisms aimed at accentuating pivotal information crucial for driver identification purposes.

\section{Methodology}\label{3}
 \begin{figure}[thpb]
  \centering
  \includegraphics[width=1.0\linewidth, height=6.7cm, keepaspectratio]{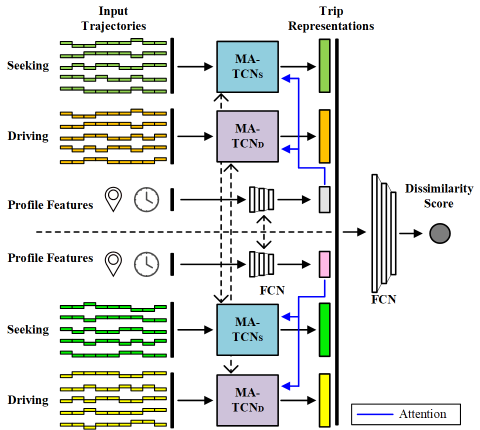}
  \vspace{-3mm}
  \caption{Siamese MA-TCN framework}
  \label{siamese MA-TCN}
\end{figure}
\begin{figure}[thpb]
  \centering
  \includegraphics[width=1.0\linewidth, height=6.7cm, keepaspectratio]{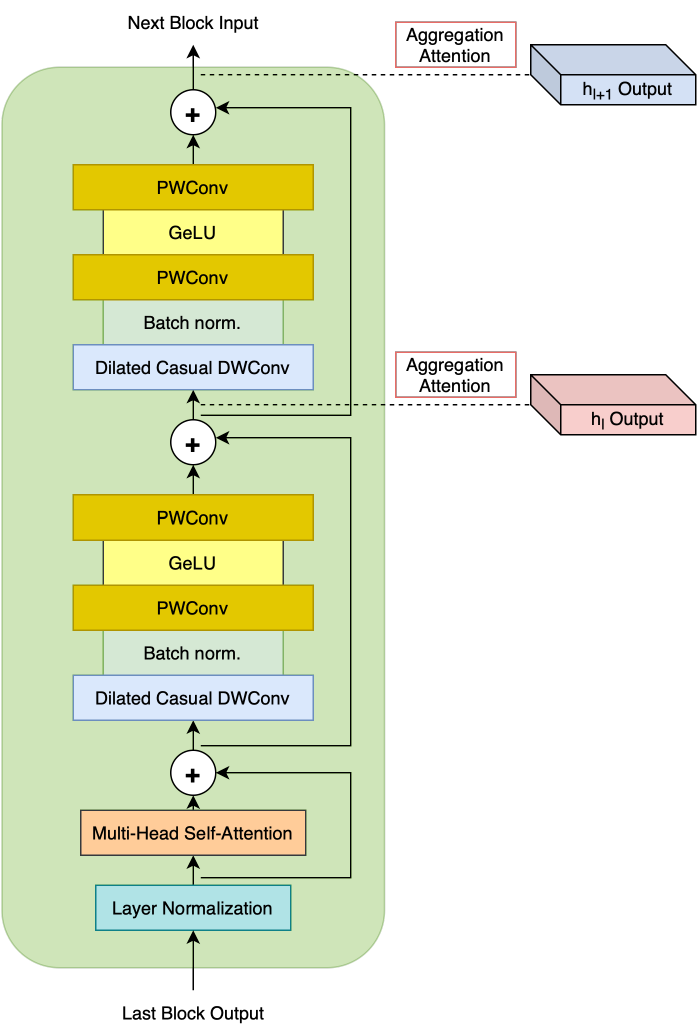}
  \vspace{-3mm}
  \caption{Multi-Head Self-Attention Double ModernTCN Block}
  \vspace{-5mm}
  \label{block}
\end{figure}
\subsection{Overview}

In this section, we briefly depict the overall architecture of the proposed Siamese Multiple Attention Temporal Convolutional Network (Siamese MA-TCN) framework for human mobility signature identification as Figure \ref{siamese MA-TCN}.

First, we use fully connected networks to learn the profile embedding $d_{emb}$ from the extracted profile features for each input driver.
Besides, for the different types (seeking or serving) of input trajectories belonging to a pair of drivers, two MA-TCN networks with identical structures and shared parameters are implemented to generate different trip representations for each type of trajectory for each driver.
Subsequently, each driver's trip representations are incorporated with the corresponding profile embedding and fed into dissimilarity learning layers composed by fully connected networks to generate the dissimilarity score.

\subsection{MHSA Double ModernTCN Block}


We present the Multi-Head Self-Attention Double ModernTCN Block (MHSA Double ModernTCN Block) as a solution for comprehensively extracting both local and long-term dependencies inherent within trajectories, as depicted in Figure \ref{block}. 
It comprises three core elements: a multi-head self-attention layer and two depthwise separable convolutional residual blocks with the same size of dilation factor.
The multi-head self-attention layer is designed to accentuate crucial time-step information within the trajectory sequence, while the depthwise separable convolutional residual blocks are tailored to capture localized features within their respective receptive fields.

\subsubsection{Multi-head Self-attention Layer}
Given the input sequence $\mathcal{I} \in \mathbb{R}^{len_{max} \times d}$, a scaled dot-product self-attention mechanism is applied as:
\begin{equation}
    \begin{aligned}
Head_i=Attention(Q, K, V)=softmax(\dfrac{QK^T}{\sqrt{d_k}})V,
    \end{aligned}
    \label{head}
\end{equation}
where $i$ stands for the index of the attention head, $i \in \{1,\dots,n\}$, n is the total number of attention heads, $Q, K, V \in \mathbb{R}^{len_{max} \times d}$ are obtained from $\mathcal{I}$ by linear transformations and $d_k=\biggl\lfloor\dfrac{d}{n}\biggr\rfloor$.
Subsequently, multiple attention heads are integrated to enhance the model's representational capacity.
By incorporating the influence of other time steps in its representation at each time step, global dependencies of trajectory sequences are captured.
The final output after passing through the multi-head self-attention layer is as :
\begin{equation}
    \begin{aligned}
    MultiHead(Q, K, V)=W_O[Head_1, \dots, Head_n] + \mathcal{I},
    \end{aligned}
    \label{multi-head}
    \end{equation}
where $W_O$ is a learnable parameter matrix, $W_O \in \mathbb{R}^{d \times d}$.


   
\subsubsection{Depthwise Separable Convolutional Residual Blocks}
To capture temporal dependencies and local features of trajectories across various granularities, thereby yielding semantically enriched representations, we employ the Temporal Convolutional Network (TCN) architecture featuring dilated convolutions to achieve an expanded receptive field size (RFS) at a reduced computational expense. 
The RFS is influenced by three parameters: the number of residual blocks (N), the convolution kernel size (K), and the dilation factor (B), defined as follows:
\begin{equation}
    \begin{aligned}
    RFS = \dfrac{2(B^N-1)(K-1)}{B-1}+1.
    \end{aligned}
    \label{RFS}
    \end{equation}

In addition, in mainstream sequence analysis tasks, existing works \cite{moderntcn}\cite{largekernel} demonstrate that larger convolutional kernels are more advantageous in capturing long-term dependencies.
We therefore further explore using larger convolutional kernels rather than more residual blocks without sacrificing the RFS when extracting the long-term dependencies on trajectory sequences.
To balance the storage and computational expense, depthwise separable convolution is adopted.
It decouples traditional convolution operation into depthwise convolution for learning temporal information and pointwise convolution for mixing information across feature channel dimensions, which is designed to greatly reduce the number of parameters, thus saving the overheads.
Besides, causal convolution makes the output of a time step rely only on elements of the current time step and earlier, which avoids being affected by padding values at the end of sequences.

 \begin{figure}[thpb]
      \centering
      \includegraphics[scale=0.53, width=1.0\linewidth, height=5.7cm, keepaspectratio]{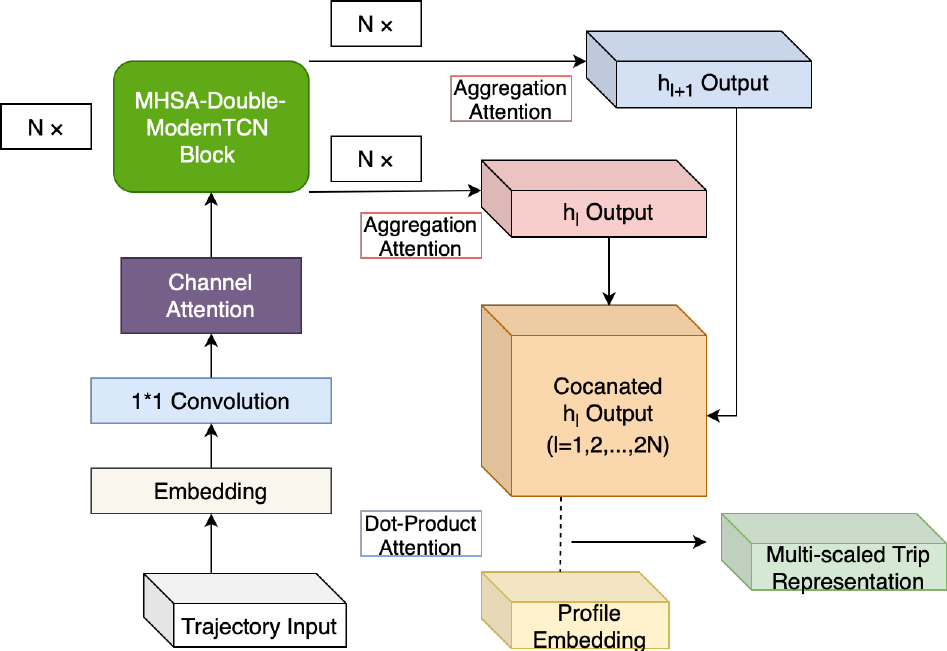}
      \caption{Multiple Attention Temporal Convolutional Network}
      \vspace{-5mm}
      \label{MA-TCN}
   \end{figure}
    
\subsection{MA-TCN Network}



Figure \ref{MA-TCN} illustrates the complete structure of the proposed MA-TCN network.
The trajectory input, represented as a sequence of grid coordinates, first passes through the embedding layer \cite{embedding} thus mapping the discrete values in the input into the embedding space and obtaining a low-dimensional real vector.
Such processing helps to discover and share similar patterns among different trajectories.
Subsequently, the embedded input tensor is successively fed into a 1×1 convolutional layer and a channel attention layer \cite{SE-net}.
The former is used to expand the size of the hidden feature channel dimensions to obtain a richer representation, and the latter is used to further emphasize the information of the key hidden feature channels by assigning different weights to different channels, thus weakening the interference of the non-key features.
To extract the local and long-term time dependencies within the trajectories comprehensively, we derive trajectory representations at different scales by stacking the proposed MHSA Double ModernTCN Blocks.

\subsubsection{Multi-scale Aggregation Attention}

To efficiently aggregate the outputs of the individual time steps obtained from each residual block, a time-wise aggregation attention mechanism is first designed.
The contribution of the output of each time step $h_{j}$, $j \in \{1,\dots,t\}$, (other than the padding time step) to the output of the last non-pad time step of each sequence $h_{t}$ (i.e., the actual last time step of each trajectory) is first considered on each residual block by assigning different attention weights to them.
The attention weight of time step \textit{j} at the \textit{l}-th residual block is:
\begin{equation}
    \begin{aligned}
    \lambda_j(l;\theta) = \dfrac{exp(\mathbf{h}_j^{(l)T}\mathcal{W}^{(l)}\mathbf{h}_t^{(l)})}{\sum_i exp(\mathbf{h}_i^{(l)T}\mathcal{W}^{(l)}\mathbf{h}_t^{(l)})}
    \end{aligned}
    \label{aggregation}
    \end{equation}
where $\mathcal{W}^{(l)}$ is a learnable parameter matrix at the \textit{l}-th residual block.
The final representation of each residual block is yielded as Equation \ref{h_l representation} by weighting and summing the information from all time steps in the time dimension.
\begin{equation}
    \begin{aligned}
\mathbf{h}_l=\sum_j\lambda_j(l;\theta)\mathbf{h}_j^{(l)}
    \end{aligned}
    \label{h_l representation}
    \end{equation}

Subsequently, to highlight representations that have a positive impact on the final result, the previously acquired profile embedding $d_{emb}$ of each driver is selected as the target using a target-specific attention mechanism:
\begin{equation}
    \begin{aligned}
    \mathbf{h}'_l=tanh(\mathbf{V}_d\mathbf{h}_l+\mathbf{b}_d),
    \end{aligned}
    \label{target1}
    \end{equation}
\begin{equation}
    \begin{aligned}
    {\beta}_l = \dfrac{exp(\mathbf{d}_{emb}^T\mathbf{h}'_l)}{\sum^{2N}_n exp(\mathbf{d}_{emb}^T\mathbf{h}'_n)},
    \end{aligned}
    \label{target2}
    \end{equation}
where \textbf{N} is the number of MHSA Double ModernTCN Blocks, and the final trip representation $Tr$ is denoted as:
\begin{equation}
    \begin{aligned}
\mathbf{Tr}=\sum^{2N}_l\beta_l\mathbf{h}_l.
    \end{aligned}
    \label{trip}
    \end{equation}
    
\subsection{Model Training}


\subsubsection{Data Preprocessing}

First, we collect the dataset based on the map data and remove the GPS sampling points that are beyond the investigated area.
For the unordered dataset, we sort the dataset by driver ID, and then group the GPS sample points according to the change of status which indicates whether there are passengers on board, and the sample points within a group are regarded as a trajectory.
In addition, trajectories containing too few (less than 10) and too many (more than 300) sampling points are excluded.
The former may be due to anomalous changes in the status leading to incorrect segmentation of the trajectory, while the latter may contain a large amount of noise.
The trajectories are then categorized into seeking and driving trajectories based on status, and drivers with less than 5 trajectories of either type in a single day are deleted because the amount of data is insufficient to explore their patterns.
Thereby, we obtain the original sequence of trajectories $\mathcal{T}$ = \{$p_{1}$, ..., $p_{m}$\}, where $p_{i}$ = [$lat_{i}$, $lon_{i}$, $t_{i}$], and $i \in \{1, \dots, m\}$, denotes the latitude and longitude of the GPS point at timestamp $t_i$.
The average velocity between each point and the next point is further calculated as its corresponding velocity $v_{i}$, thus updating the sequence of trajectories as $\mathcal{T'}$ = \{[$p_{1}$, $v_{1}$]\, ..., [$p_{m}$, $v_{m}$]\}.

To minimize computational overhead, the investigated area is subsequently divided into equal-sized grid cells with a given side length $s$ in latitude and longitude ($s$ = 0.01°).
Besides, each day is divided into 5-minute intervals for a total of 288 intervals per day, denoted as $T$ = {$interval_{k}$}, where $1 \leq k \leq 288$.
Based on these operations, the sequence of trajectories $\mathcal{T'}$ can then be represented by a sequence of grid coordinates $\mathcal{G}$ = \{[$g_{lat1}$, $g_{lon1}$, $interval_{1}$, $v_{1}$], ..., [$g_{latm}$, $g_{lonm}$, $interval_{m}$, $v_{m}$]\}, where $g_{lati}$ stands for the coordinates in the latitude direction, $g_{loni} $ represents the coordinates in the longitude direction in the grid coordinate system, $i \in \{1, \dots, m\}$.
Since the lengths of the trajectories are not equal to each other, for better parallelism, each trajectory is padded to the maximum length of the trajectories $len_{max}$ in each training batch to get the equal-length sequence $G_{pad} \in \mathbb{R}^{len_{max} \times 4}$.
In addition, the padded trajectories are used to generate a padding mask to label the actual length of each trajectory, thus eliminating the effect of the end padding values on the results during the subsequent attention computation.
Recall that we transformed the feature dimensions of $G_{pad}$ by embedding layers and the 1×1 convolutional layer, so that eventually, before entering MHSA Double ModernTCN Block, the trajectory sequence is represented as $\mathcal{I} \in \mathbb{R}^{len_{max} \times d}$.

\subsubsection{Loss Function}

During the training phase, the binary cross entropy loss is employed as the criterion, aimed at minimizing the dissimilarity metric between trajectories originating from the same driver while maximizing it for those originating from distinct drivers, shown as
\begin{equation}
    \begin{aligned}
\min_\theta -(ylog(D_\theta(X_1, X_2))+(1-y)log(1-D_\theta(X_1, X_2)))\\
\text{s.t.} X_1=(Tr_{s, 1}, Tr_{d, 1}, d_1),
            X_2=(Tr_{s, 2}, Tr_{d, 2}, d_2),
    \end{aligned}
\end{equation}
where it holds that $y=0$ if the trajectories belong to the same driver and $y=1$ if the trajectories are sampled from two different drivers.
$D_\theta(X_1, X_2)$ is the prediction probability of how likely the trajectories belong to the two different drivers.

\section{Experiment}\label{5}

\subsection{Experiment Setting}
\subsubsection{Data Description}

In this section, we evaluate our model on two large-scale datasets:
\begin{itemize}
    \item Shenzhen Dataset: The raw Shenzhen dataset contains GPS records collected from taxis in Shenzhen, China during July 2016. There were in total 17,877 taxis equipped with GPS sets, where each GPS set generates a GPS point every 40 seconds on average.
    We conduct the experiments on the dataset provided in \cite{st_siamese}, which contains a total of 158,718 trajectories from 697 drivers over 10 workdays, with each trajectory containing an average of about 22 sampled GPS points.
    \item Chengdu Dataset: The raw Chengdu dataset is collected from real-world taxis in Chengdu, China dated from Aug 3rd, 2014 to Aug 29th, 2014. Over 1.4 billion GPS records are collected and over 14,000 taxis are involved \cite{chengdu_dataset}.
    We select a total of 585,483 trajectories of 697 drivers over a complete two-week period, (i.e., 14 days, from Aug 3rd, 2014 to Aug 16th, 2014), for the experiments.
    Each trajectory contains an average of about 60 sampled GPS points.
\end{itemize}
Each GPS sample point in both datasets contains five key data fields, including taxi ID, time stamp, status, latitude, and longitude. 
Status is a binary value that indicates whether there are passengers on board.
For the Shenzhen/Chengdu dataset, we use the data from the first 5/7 days of the 500 drivers for training and the remaining 5/7 days of data from these drivers for validation.
Subsequently, the remaining 197 drivers who were not involved in the training were tested using the trajectory data of the latter 5/7 days.

\subsubsection{Hyperparameter Setting}    
The Hyperparameters we used in our experiment are described as follows:
\begin{itemize}
\item We set the latent feature dimension $d$ to 64.
Besides, the hidden size of the multi-head self-attention is also set to 64 and there are 8 attention heads in total.
\item We set the number of MHSA Double ModernTCN Block $N$ to 4, the kernel size of depthwise convolution $K$ to 10, and the dilation factor $B$ to 2.
\end{itemize}
We implemented our model using PyTorch. 
The model is trained using Adam optimizer on the server with 2.1 GHz Intel Xeon E5-2620 v4 CPUs, 128G RAM, and nVidia 2080Ti 11G GPUs.
The initial learning rates are 0.0001 and 0.00006 for the Shenzhen and Chengdu datasets, respectively.

\begin{table}[h]
  \centering
  \caption{Performance on Shenzhen dataset comparing the baselines}
  \label{shenzhen}
  \begin{tabular}{lccc}
  \toprule
     \multirow{2}{*}{\textbf{Methods}} & \multicolumn{3}{c}{\textbf{Shenzhen}}  \\ 
     \cmidrule{2-4} 
    &     Accuracy    & Recall     & $F_{1}$ Score    \\ \midrule
                 Siamese MA-TCN &                          0.870 & \textbf{0.8641}            & \textbf{0.8623}     \\
     ST-SiameseNet                 & \textbf{0.871} & 0.8317 & 0.8508  \\
                            LSTM  & 0.847                         & 0.8130 & 0.835   \\
                           FNN   & 0.6112             & 0.6298 & 0.6195  \\ 
                           SVM  & 0.81             & 0.7661 & 0.7874   \\ 
  \bottomrule
  \end{tabular}
\end{table}
\begin{table}[h]
  \centering
  \caption{Performance on Chengdu dataset comparing the baselines}
  \label{chengdu}

  \begin{tabular}{lccc}
  \toprule
     \multirow{2}{*}{\textbf{Methods}} & \multicolumn{3}{c}{\textbf{Chengdu}}  \\ 
     \cmidrule{2-4}  
    &     Accuracy    & Recall     & $F_{1}$ Score         \\ \midrule
                 Siamese MA-TCN &                           \textbf{0.790}         & 0.5991       & \textbf{0.7193 } \\
     ST-SiameseNet                 &  0.759 & 0.5662  &  0.6976   \\
                            LSTM  & 0.744 & 0.5220 & 0.6605   \\
                           FNN   &  0.693 & 0.5340 & 0.6483 \\ 
                           SVM  &  0.681 & \textbf{0.7213} & 0.6881 \\ 
  \bottomrule
  \end{tabular}
  \vspace{-3mm}
\end{table}

\subsection{Performance Comparison}
\subsubsection{Evaluation Metrics}
In order to assess the efficacy of our proposed model alongside baseline methods, we conduct evaluations based on accuracy, recall, and F1 score, compared against ground truth labels. 
In our implementation, a dissimilarity score threshold of 0.5 was employed.
Trajectories with scores below this threshold were deemed to originate from the same driver, while those surpassing it were attributed to different agents. 
Accuracy, gauges the classifier's proficiency in distinguishing between distinct agents, while recall measures its capacity to avoid overlooking pairs of dissimilar drivers. 
The F1 score, as a harmonic mean of precision and recall, offers a balanced assessment, amalgamating the model's precision and recall metrics.

\subsubsection{Baseline Methods}
We compare our model with the following baseline methods:
\begin{itemize}
\item Support Vector Machine (SVM): We use the set of profile features as input, calculate the absolute error of the profile features of two drivers, and then identify whether these two drivers are the same or not using a support vector machine.
\item Fully Connected Neural Network (FNN): Fully Connected Neural Network is the basic classification or regression model in deep learning.
All the trajectories of two drivers and their profile features in two time periods are connected together as inputs to the neural network.
\item Long Short-Term Memory Network (LSTM): Instead of using the architecture of the siamese network, a separate LSTM is trained for each type of trajectory for each driver.
A total of four LSTM networks are trained, and the other settings are the same as for ST-SiameseNet.
\item ST-SiameseNet: A siamese network for HuMID proposed by Ren \cite{st_siamese} in 2020.
Two siamese LSTM networks are utilized for trajectory feature extraction.
\end{itemize}

\subsubsection{Comparison Results}

The results on two trajectory datasets are shown in Tables \ref{shenzhen} and \ref{chengdu}.
It is clear that our model has the best overall performance. 
On the Shenzhen dataset, compared to the state-of-the-art baseline method ST-SiameseNet, our model has the higher recall and $F_{1}$ score with close accuracy under the same dissimilarity score threshold of ST-SiameseNet.
This may be due to the fact that shorter trajectories do not have enough information at coarser granularity to fully utilize the information at different scales.
Shifting to the performance on the Chengdu dataset, compared to other baseline methods, our method achieves the highest accuracy and $F_{1}$ score.
Although the recall value is lower than that of traditional machine learning methods, it is still better than that of other methods based on deep learning, including ST-SiameseNet.
This may be due to the fact that SVMs that only use manually extracted features are less capable of extracting similarities between drivers and thus are more inclined to make dissimilarity predictions.

It is worth mentioning that the time to train our model to reach convergence on the Shenzhen dataset is reduced by 75\% compared to ST-SiameseNet. 
In addition, due to the long training time of ST-SiameseNet on the Chengdu dataset, we show the best results of ST-SiameseNet at the same time required for our model to train to convergence.

\subsubsection{Effectiveness of Different Attention Mechanisms}
In order to verify the effectiveness of our attention components, we designed the following model based on MA-TCN:
\begin{itemize}
    \item Siamese MA-TCN$_\text{MHSA}$: We remove multi-head self-attention layers and use only depth-separable residual blocks for feature extraction.
    The rest of the model design is the same as Siamese MA-TCN.
    \item Siamese MA-TCN$_\text{agg}$: We remove the multi-scale aggregation attention mechanism and use the last non-pad time step $h_t$ of the last block as the final representation of the trajectory.
    The rest of the model design is the same as Siamese MA-TCN.
\end{itemize}
and the ablation test is carried out on the Shenzhen dataset, the result is shown in Table \ref{ablation}.
\begin{table}
  \centering
  \caption{Ablation test on Shenzhen dataset}
  \label{ablation}
  \begin{tabular}{lccc}
  \toprule
     \textbf{Methods} & Accuracy    & Recall     & $F_{1}$ Score  \\ 
     \midrule
                 Siamese MA-TCN &                          \textbf{0.870} & \textbf{0.8641}            & \textbf{0.8623}     \\
                 Siamese MA-TCN$_{MHSA}$ &                          0.830 & 0.8554            & 0.8337     \\
                   Siamese MA-TCN$_{agg}$ &                          0.837 & 0.8475            & 0.8434   \\
  \bottomrule
  \end{tabular}
          \vspace{-5mm}
\end{table}
The results show that eliminating multi-head self-attention layers and aggregation attention causes a 4\% and 3.3\% loss of accuracy respectively.
Besides, the values of recall and $F_{1}$ score decrease accordingly, since the ability to represent trajectories is weakened by the lack of comprehensive consideration of local and global long-term dependencies.
Thus, the result indicates that our model integrating multi-head self-attention and aggregation attention mechanisms is capable of effective extraction of both long-term dependencies and local features.

\section{Conclusion}\label{6}

In this paper, we proposed a Siamese Multiple Attention Temporal Convolutional Network (Siamese MA-TCN) to solve the trajectory identification tasks for pairs of drivers (i.e., Human Mobility Signature Identification).
Our model integrates the benefits of the TCN architecture with a multi-head self-attention mechanism to effectively capture multi-scale local features alongside long-term dependencies, facilitated by a specially designed aggregation attention mechanism.
We conduct extensive experiments on two large-scale real-world datasets and the results show that our model achieves an efficient balance between performance and computational overhead.

\bibliographystyle{IEEEtran}
\bibliography{Reference}
\end{document}